\documentclass[conference]{IEEEtran}
\IEEEoverridecommandlockouts

\usepackage[noadjust]{cite}
\usepackage{amsmath,amssymb,amsfonts}
\usepackage{algorithmic}
\usepackage{graphicx}
\usepackage{textcomp}
\usepackage{xcolor}
\usepackage{multirow}
\def\BibTeX{{\rm B\kern-.05em{\sc i\kern-.025em b}\kern-.08em
    T\kern-.1667em\lower.7ex\hbox{E}\kern-.125emX}}
\begin{document}

\title{Concept-Guided Chain-of-Thought Prompting for Pairwise Comparison Scoring of Texts with Large Language Models
\thanks{Our code is available at \texttt{https://github.com/SMAPPNYU/CGCoT}.}
}

\author{\IEEEauthorblockN{Patrick Y. Wu}
\IEEEauthorblockA{\textit{Department of Computer Science} \\
\textit{American University}\\
Washington, DC, United States \\
patrickwu@american.edu}
\and
\IEEEauthorblockN{Jonathan Nagler}
\IEEEauthorblockA{\textit{Center for Social Media and Politics, Department of Politics} \\
\textit{New York University}\\
New York, NY, United States \\
jonathan.nagler@nyu.edu}
\and
\IEEEauthorblockN{Joshua A. Tucker}
\IEEEauthorblockA{\textit{Center for Social Media and Politics, Department of Politics} \\
\textit{New York University}\\
New York, NY, United States \\
joshua.tucker@nyu.edu}
\and
\IEEEauthorblockN{Solomon Messing}
\IEEEauthorblockA{\textit{Center for Social Media and Politics} \\
\textit{New York University}\\
New York, NY, United States \\
solomon.messing@nyu.edu}
}

\maketitle

\begin{abstract}
Existing text scoring methods require a large corpus, struggle with short texts, or require hand-labeled data. We develop a text scoring framework that leverages generative large language models (LLMs) to (1) set texts against the backdrop of information from the near-totality of the web and digitized media, and (2) effectively transform pairwise text comparisons from a reasoning problem to a pattern recognition task. Our approach, concept-guided chain-of-thought (CGCoT), utilizes a chain of researcher-designed prompts with an LLM to generate a concept-specific breakdown for each text, akin to guidance provided to human coders. We then pairwise compare breakdowns using an LLM and aggregate answers into a score using a probability model. We apply this approach to better understand speech reflecting aversion to specific political parties on Twitter, a topic that has commanded increasing interest because of its potential contributions to democratic backsliding. We achieve stronger correlations with human judgments than widely used unsupervised text scoring methods like Wordfish. In a supervised setting, besides a small pilot dataset to develop CGCoT prompts, our measures require no additional hand-labeled data and produce predictions on par with RoBERTa-Large fine-tuned on thousands of hand-labeled tweets. This project showcases the potential of combining human expertise and LLMs for scoring tasks.
\end{abstract}


\section{Introduction}
Text scoring methods are used to analyze the latent positions of texts, such as tweets, documents, or speeches, along one or more dimensions. Popular text scoring methods, such as Wordfish, are used extensively in the social and political sciences (see, e.g., \cite{laver_benoit_garry_2003,wordfish_slapin_proksch,wu_etal_2019,benoit_munger_spirling_2019,rheault_cochrane_2020,bailey2023measuring}). However, these approaches require a large text corpus, struggle with short texts, and do not have a mechanism to precisely target the latent concept of interest. Other approaches, such as fine-tuning pre-trained language models such as RoBERTa \cite{liu2019roberta}, require non-trivial quantities of hand-labeled data. These methods often fail with texts such as social media posts: they are short, it is often unclear how to pick documents for identification or labeling, and the usages of words rapidly shift over time.

In this article, we present a novel text scoring framework that leverages the embedded information and pattern recognition capabilities of generative large language models (LLMs). LLMs can set the texts against the backdrop of information accumulated from the near-totality of the web and digitized media, giving greater context to short texts such as social media posts. The core idea is to use an LLM to conduct pairwise comparisons between two texts. In other words, we prompt an LLM to pick the text that reflects a greater quantity of some latent or abstract concept of interest, such as which text contains greater aversion to a particular political party. But instead of directly pairwise comparing texts, we compare concept-specific breakdowns of the texts. These concept-specific breakdowns are generated using an approach we call concept-guided chain-of-thought (CGCoT) prompting. CGCoT prompting is a framework that uses a series of researcher-crafted questions that examine the constituent parts of the concept of interest in a given text. The text and the LLM's answers to the CGCoT prompts for that text form the text's concept-specific breakdown. These researcher-crafted prompts, akin to a codebook used to guide content analysis, are the same across all texts, making the concept-specific breakdowns directly comparable in pairwise comparisons.

We use the LLM to pairwise compare concept-specific breakdowns of the texts along a targeted concept. Following this, we score the LLM's answers using the Bradley-Terry model, a probabilistic model that predicts the outcome of pairwise comparisons based on the latent abilities of the items being compared \cite{bradleyterry1952}. We call the resulting ``ability'' scores of the texts CGCoT pairwise scores. Because we craft the CGCoT prompts and the basis of the pairwise comparisons, we can precisely target the latent concept of interest. We rely on pairwise comparisons rather than attempting to use the LLM to directly adjudicate the level, intensity, or scalar value of concepts like aversion for a number of reasons. First, pairwise comparisons enable us to establish a measure where differences are meaningful; ordinal rankings indicate order without quantifying the gaps between ranked items. Second, scalar values directly generated from the LLM lack transparency in their generation and calibration. In contrast, the Bradley-Terry model derives estimated scalar values from pairwise comparisons, providing a more interpretable and reliable basis for the score. Third, pairwise comparisons are easier to complete and allow for more subtle distinctions, improving reliability over labeling tasks over single items \cite{carlson_montgomery_2017}.

We apply the proposed approach to better understand affective polarization on Twitter \cite{yarchi_2021,nordbrandt_2021}. Affective polarization is the tendency for partisans to dislike or distrust members of the opposing party \cite{iyengar_sood_lelkes_affectivepolarization,Druckman2021}. Affective polarization is typically studied using surveys but, except for \cite{affective_polarization_csmap}, has not been extensively studied in the context of political non-elites on social media. The authors in \cite{affective_polarization_csmap} label tweets using a dichotomous classification for aversion to Republicans and then again for aversion to Democrats in tweets. Expressions of aversion to a specific party are an inevitability on social media, but strong expressions of aversion threaten productive discourse online, incentivize antidemocratic rhetoric, and, when analyzed in the aggregate, can signal the declining health of democracy \cite{finkel_etal_2020_sectarianism}. 

We calculate two aversion scores using a random sample of political tweets from \cite{affective_polarization_csmap}: an aversion to Republicans score and an aversion to Democrats score. We develop a series of questions used as CGCoT prompts with the LLM---here, we use GPT-3.5---that identify and describe aversion to specific parties in a given tweet. Specifically, we use three prompts: the first prompt uses GPT to summarize the tweet; the second prompt uses GPT to identify the primary party that is the focus or target of the tweet; the third prompt uses GPT to identify whether aversion is expressed towards the targeted party. We apply these prompts to each tweet in the corpus: the tweet's concept-specific breakdown is the original tweet and all the LLM's responses to the CGCoT prompts about the tweet. Then, using a sample of pairwise comparisons between the concept-specific breakdowns, we prompt GPT-3.5 to select the breakdown that exhibits greater aversion to a specific party. Lastly, using the outcomes of these pairwise comparisons, we estimate an aversion score for each text using the Bradley-Terry model.

To validate the CGCoT pairwise scores, we compare CGCoT pairwise scores with three alternative unsupervised text scoring approaches. We show that pairwise comparing the concept-specific tweet breakdowns yields scores that are more strongly associated with human judgments than the three alternative text scoring approaches. The results indicate that using both CGCoT prompts and pairwise comparisons is important for deriving high-quality scores. We also show that our continuous score, which only requires a small set of pilot hand-labeled tweets to develop the CGCoT prompts, is competitive with state-of-the-art supervised approaches. Binarizing the scores---classifying all tweets with a CGCoT pairwise score above a cutoff as exhibiting aversion and all observations with a score below a cutoff as not exhibiting aversion---gives us a set of predictions that nearly match (for aversion to Democrats) or exceed (for aversion to Republicans) the performance of a RoBERTa-Large \cite{liu2019roberta} model fine-tuned on 3,000 hand-labeled tweets. Overall, the success of CGCoT in measuring aversion suggests that pairing substantive knowledge with LLMs can be immensely useful for solving social science text measurement problems. 

\section{Related Work}
Our approach is situated in a rapidly growing literature on using generative LLMs for social science applications \cite{tornberg2023chatgpt4,rathje_mirea_sucholutsky_marjieh_robertson_vanbavel_2023,argyle2023ai,bisbee_clinton_dorff_kenkel_larson_2023,wu2023large}. The works that study text, such as analyzing text along psychological constructs \cite{rathje_mirea_sucholutsky_marjieh_robertson_vanbavel_2023}, analyze the text as given. Our proposed text scoring approach breaks the text down into the concept of interest's constituent parts using prompts developed by substantive knowledge about the targeted latent concept; it involves substantive expert knowledge to a much greater extent than the other research studies that analyze text using LLMs.

This framework also speaks to a large body of text scoring methods. These text scoring methods roughly fall into unsupervised methods and supervised methods. Unsupervised methods such as Wordfish \cite{wordfish_slapin_proksch} and word embedding methods \cite{kozlowski_taddy_evans_2019,wu_etal_2019,an-etal-2018-semaxis,frameaxis} typically require post hoc dimensional interpretation or selection of keywords to represent underlying concepts of interest and a large corpus. Supervised methods, such as WordScores \cite{laver_benoit_garry_2003,lowe_2017} and approaches that use pairwise comparisons of texts \cite{LOEWEN2012212,simpson-etal-2019-predicting}, rely on hand-labeled texts or manual pairwise comparisons of texts and typically focus on measuring one targeted concept within the corpus. Our approach minimizes the need for hand-labeling or manual pairwise comparisons of texts, can measure multiple targeted concepts within the same corpora, does not require a large corpus, relies on a transformers-based language model rather than bag-of-words, and leverages the researchers' substantive knowledge to precisely target the latent concept of interest instead of relying on post hoc dimensional interpretation.

Many recent works have also shown that generative LLMs can outperform crowd workers for text-annotation tasks \cite{gilardi_etal_2023,tornberg2023chatgpt4}. These papers usually focus on discrete classification tasks. The highly structured nature of these tasks, with clear gold standard comparisons, plays to the strengths of LLMs. However, it is less clear if these advantages hold with continuous, open-ended, and contentious concepts such as aversion to opposing parties. Some works have also directly queried scalar values from the LLM \cite{ohagan2023measurement}. However, LLMs are not inherently designed to produce consistent and calibrated numbers. The generated scalar values may change with different prompts, training updates, and so on. Pairwise comparisons, on the other hand, are a task that is more aligned with the LLM's training: determining which of two items or texts have a greater quality is similar to other natural language tasks such as entailment, and is more compatible with the LLM's strengths in natural language tasks.

In our proposed framework, pairwise comparisons are made over \textit{concept-specific breakdowns} of the texts rather than the texts themselves. It is well-documented that generative LLMs often make mistakes in problems that require intermediate reasoning steps \cite{liu-etal-2022-generated,wei2023chainofthought,kojima2023large,zhou2023leasttomost,wu2023reasoning}. The authors in \cite{wei2023chainofthought} propose ``chain-of-thought,'' which prompts the LLM to explicitly generate its intermediate reasoning steps, leading to improved responses on problem-solving tasks such as arithmetic or question-answering. In related work, the authors in \cite{zhou2023leasttomost} use the LLM to automatically break a problem down into subproblems using few-shot examples, an approach they call least-to-most prompting; the LLM then solves each subproblem to solve a larger, harder problem. But despite these innovations, it is still unclear whether generative LLMs are able to ``reason.'' For instance, the authors in \cite{wu2023reasoning} find that LLMs perform poorly on ``counterfactual'' tasks, which are variants of reasoning tasks that LLMs performs well on.

Political polarization on social media has also been studied using natural language processing techniques. Many of these studies focus on the framing devices used in social media posts \cite{demszky-etal-2019-analyzing,grover-etal-2019}, the discovery of political polarization on social media \cite{belcastro-etal-2020}, or the examination of similar language used to express opposing viewpoints \cite{KhudaBukhsh_Sarkar_Kamlet_Mitchell_2021}. Our work differs in that it analyzes affective polarization, a type of polarization based on social identity rather than policy-based division \cite{iyengar_sood_lelkes_affectivepolarization}. We also estimate measures of specific dimensions of affective polarization, rather than discovering framing devices or the existence or extent of polarization.

\section{The Text Pairwise Comparison Framework using CGCoT}
We build on previous chain-of-thought work by proposing CGCoT prompting. Rather than using the LLM to generate its own intermediate reasoning steps or its own breakdown of the problem into subproblems, we leverage the researcher's substantive knowledge about the targeted latent concept to craft a sequence of questions that identify and describe the targeted concept and its constituent parts in the text. This approach is analogous to using a codebook for qualitative content analysis \cite{FONTEYN2008165}. In other words, we use the LLM's pattern recognition capabilities in conjunction with researcher-guided prompts to generate the intermediate reasoning steps that we would ideally like the LLM to reason through for a targeted concept when making pairwise comparisons. CGCoT effectively shifts the pairwise comparisons of text from a reasoning problem to a pattern recognition task. For example, if we are scoring the level of aversion expressed towards a target in a text, we can use a series of prompts that summarizes the text, identifies the primary focus or target of the text, and identifies whether aversion is expressed towards that target.

To be more precise, generating the concept-specific breakdowns follows these steps:
\begin{enumerate}
    \item Let $t_i$ be a text, for some $i \in \{1,...,n\}$, and let $(x_1,...,x_m)$ be a set of $m$ researcher-crafted concept-guided prompts to extract specific information from $t_i$
    \item Sample a token sequence using an LLM with parameters $\theta$, $s_{1,i} \sim p_{\theta}\left(s | x_1, t_i\right)$
    \item Then, sample a token sequence $s_{2,i} \sim p_{\theta}\left(s | x_2, t_i, s_{1,i}, x_1\right)$
    \item Repeat this iterative sampling approach with all prompts, such that the last token sequence sampled is $s_{m,i} \sim p_{\theta}\left(s | x_m, t_i, s_{m-1,i}, x_{m-1}, s_{m-2,i}, x_{m-2}, ..., s_{1,i}, x_1\right)$
    \item Concatenate all sampled token sequences $s_{j,i}$ for $j \in \{1,...,m\}$ and text $t_i$ to form the concept-specific breakdown
    \item For concept-guided prompt development, compare the concept-specific breakdown with a set of hand-labeled text data to assess if concepts and entities are correctly identified; if not, refine prompts $(x_1,...,x_m)$
\end{enumerate}

After generating the concept-specific breakdowns for each text in the corpus, we pairwise compare the breakdowns instead of the texts themselves. The pairwise comparison prompt is determined based on the application. We then use the outcomes of the pairwise comparisons with the Bradley-Terry model to estimate a scalar score for each text.

\subsection{Bradley-Terry Model}
The Bradley-Terry model assumes that in a contest between two players $i$ and $j$, the odds that $i$ beats $j$ in a matchup are $\alpha_i / \alpha_j$, where $\alpha_i$ and $\alpha_j$ are positive-valued parameters that indicate latent ``ability'' \cite{bradleyterry1952}. We can define $\alpha_i \equiv \exp(\lambda_i)$. Then, the log-odds of $i$ beating $j$ is 
\[\log\left[ \frac{\text{Pr}(i \text{ beats } j)}{\text{Pr}(j \text{ beats } i)} \right] = \lambda_i - \lambda_j\]
The intuition is that the larger the value of $\lambda_i$ compared to $\lambda_j$, the more likely it is for player $i$ to beat player $j$. 

We translate the above matchup into a contest between two concept-specific breakdowns. Using the aversion measures as our example, the estimated $\lambda$ parameters are the measures of the level of aversion to a specific party. We denote the concept-specific breakdown exhibiting greater aversion to a specific party as the ``winner.'' We considered ties as 0.5 wins for both tweets in the matchup. The authors in \cite{bt_turner_firth} find that this approach yields ability parameter estimates that highly correlate with more complex approaches that explicitly deal with ties. We use the bias-reduced maximum likelihood estimation approach implemented in the \texttt{BradleyTerry2} R package with GPT's answers to pairwise comparisons to estimate the level of aversion expressed towards a specific party in each tweet. These scores are the aversion CGCoT pairwise scores. The estimated $\lambda$ parameters are relative to a reference tweet, but this choice is unimportant because we rescale the parameters to the unit interval. 

We also estimate standard errors for the estimated $\lambda$ parameters. These standard errors are interpreted relative to a reference tweet. We calculate quasi-variances from these standard errors, which can be interpreted as reference-free estimates of the variance of the score of each tweet. Confidence intervals derived from these quasi-variances can be directly compared. We use the \texttt{qvcalc} package to calculate quasi-standard errors \cite{qvcalcfirth}. The 95\% confidence intervals of the estimated scores are derived from these quasi-standard errors.

\section{Description and Prompts Used for Application: Analyzing Aversion to Opposing Parties on Social Media}
\subsection{Data}
We use GPT-3.5, with its default temperature and nucleus sampling hyperparameter values, to pairwise compare political tweets from \cite{affective_polarization_csmap}. The authors in \cite{affective_polarization_csmap} use 3,000 coder-labeled tweets to fine-tune a RoBERTa model that classifies tweets as containing aversion to Republicans and/or aversion to Democrats in a multilabel setting. 500 tweets are used for validation, and 500 tweets are set aside as a test set. These tweets were selected using a set of political keywords from the Twitter Decahose. Each tweet was labeled by 3 coders from Surge AI. 33.9\% of the tweets were labeled as expressing aversion to Republicans, and 31.2\% of the tweets were labeled as expressing aversion to Democrats. The average Cohen's $\kappa$ was 0.795 for aversion to Republicans and 0.794 for aversion to Democrats. Using the approach outlined in the previous section, we score the test set tweets to make our results comparable with the results from \cite{affective_polarization_csmap}.

These tweets were coded by Surge AI in 2023, and neither the corpus of tweets nor the labels have been posted online. Therefore, this dataset is not part of GPT-3.5's (nor any LLM's) training corpus. 

\subsection{Prompt to Generate Concept-Specific Breakdowns}
To generate aversion to Republican-specific breakdowns, we created the following concept-guided prompts using definitions and concepts from the literature on affective polarization \cite{iyengar_sood_lelkes_affectivepolarization,finkel_etal_2020_sectarianism,Druckman2021,affective_polarization_csmap}:

\begin{enumerate}
\begin{ttfamily}
    \small
    \item Summarize the Tweet. 
    \item We broadly define Republicans to include any member of the Republican Party/GOP, the Republican Party/GOP generally, conservatives, right-wingers, anyone that supports MAGA, or the alt-right. We broadly define Democrats to include any member of the Democratic Party, the Democratic Party generally, liberals, leftists, or progressives. Using these definitions, does the Tweet primarily focus on Republicans (or a Republican) or Democrats (or a Democrat)? The focus can be on a specific member of a party.
    \item If the Tweet primarily focuses on Republicans based on your above answer, does the Tweet express aversion, dislike, distrust, blame, criticism, or negative sentiments of Republicans (or a Republican)? If the Tweet primarily focuses on Democrats based on your above answer, does the Tweet express aversion, dislike, distrust, blame, criticism, or negative sentiments of Democrats (or a Democrat)? If the Tweet focuses on neither party, answer ``N/A.''
    \item Using only your answer immediately above, does the Tweet express aversion, dislike, distrust, blame, criticism, or negative sentiments of Republicans (or a Republican)?
\end{ttfamily}
\end{enumerate}

We used a similar set of prompts to create aversion to Democrats-specific breakdowns: the first three questions are the same, except we flipped the order of the definitions, and we replaced the word ``Republicans'' with ``Democrats'' in the fourth question. Flipping the order of the definitions controls for potential order effects in the prompts used to generate the breakdown for each aversion to a specific party. The last prompt provides information on which party is the subject of any aversion. Taking advantage of the conversational aspect of generative LLMs, and as detailed in the previous section, we prompt each question sequentially. The concept-specific breakdown is the concatenation of the original tweet and the LLM's responses to all four prompts.

The process of developing these prompts is analogous to creating a codebook for human coders and qualitative content analysis \cite{satu_kyngas_2008,FONTEYN2008165}. To ``make sense of the data and whole,'' we labeled 50 tweets as containing aversion to Republicans, 50 tweets as containing no aversion to Republicans, 50 tweets as containing aversion to Democrats, and 50 tweets as containing no aversion to Democrats from \cite{affective_polarization_csmap}'s training set; we did not use their labels \cite{satu_kyngas_2008}. We then iterated on an initial set of CGCoT prompts and examined outputs from GPT-3.5 until the summaries and party identifications aligned with expectations across labeled tweets. In short, we combined content analysis techniques with prompt engineering methods such as changing specific words, repeatedly providing definitions, and splitting up complex questions.

\subsection{Pairwise Comparing Concept-Specific Breakdowns}
\label{subsection:pairwise_comparison}
To pairwise compare these concept-specific breakdowns generated using the prompts from the previous section, we prompt GPT-3.5 to pick the breakdown that expresses greater aversion to a specific party. We pairwise compare the concept-specific breakdowns for aversion to Republicans using the following prompt:

\begin{quote}
\begin{ttfamily}
    \small
    Tweet Description 1: [concept-specific breakdown for the first tweet]
    
    Tweet Description 2: [concept-specific breakdown for the second tweet]
    
    Based on these two Tweet Descriptions, which Tweet Description expresses greater aversion, dislike, distrust, blame, criticism, or negative sentiments of Republicans: Tweet Description 1 or Tweet Description 2? If both equally express or do not express aversion, distrust, blame, criticism, or negative sentiments of Republicans, reply with ``Neither'' or ``Tie.''
\end{ttfamily}
\end{quote}

We use this same prompt for aversion to Democrats, except we replace the word ``Republicans'' with ``Democrats.'' When we compare tweets directly in Section~\ref{sec:comparing_cgcot_human_coders}, we also use the same prompt except that we replace all instances of ``Tweet Description'' with ``Tweet'' and all instances of ``Tweet Descriptions'' with ``Tweets.'' 

For each pairwise comparison, GPT-3.5 typically outputs a paragraph explaining its choice. Instead of restricting GPT-3.5's answers to only ``Tweet Description 1'' or ``Tweet Description 2,'' we use a separate prompt to extract the answers. We find that this two-step process improves GPT's responses in pairwise comparisons. To extract the answer from the response, we first concatenate the model's response to the pairwise comparison with the following text:

\begin{quote}
\begin{ttfamily}
    \small
    In the above Text, which Tweet Description is described to be expressing greater aversion, dislike, distrust, blame, criticism, or negative sentiments of Republicans: Tweet Description 1 or Tweet Description 2? Return only ``Tweet Description 1'' or ``Tweet Description 2''. If neither Tweet Descriptions are described to be more likely to be expressing greater aversion, dislike, distrust, blame, criticism, or negative sentiments of Republicans, reply with ``Tie.''
\end{ttfamily}
\end{quote}

This is then used as a prompt for GPT-3.5. We manually fixed a very small number of answers that deviate from ``Tweet Description 1,'' ``Tweet Description 2,'' or ``Tie.'' Again, we use the same prompt for extracting answers from non-CGCoT tweets-only comparisons in Section~\ref{sec:comparing_cgcot_human_coders}, except we replace the ``Tweet Description'' with just ``Tweet.'' 

There are a total of 124,750 potential matchups. To reduce the total number of matchups, we sample 20 matchups per tweet ID for a total of 10,000 matchups. The Bradley-Terry model does not require complete matchups to estimate scores for each tweet. In the Appendix, we show that CGCoT pairwise scores estimated using 5, 10, and 15 matchups per tweet ID highly correlate for both aversion scores. The scores are then rescaled to a 0-1 range, making each score independent of any reference tweet. We also estimate 95\% confidence intervals based on quasi-standard errors.

\subsection{Example of Applying CGCoT on Two Tweets}
We walk through an example of one pairwise comparison using concept-specific breakdowns. The first tweet reads: ``He's such a lowlife piece of shit. Isn't this slander? Can he be sued? And the GOP are silent. What crap they are. They did this. They own this. \#NeverForget \#VoteBlueToSaveAmerica.'' The concept-specific breakdown for this tweet, using the CGCoT prompts for aversion to Republicans, is the following.

\begin{quote}
\begin{ttfamily}
\small
(0) Original Tweet: ``He's such a lowlife piece of shit. Isn't this slander? Can he be sued? And the GOP are silent. What crap they are. They did this. They own this. \#NeverForget \#VoteBlueToSaveAmerica''

(1) The tweet contains a link and expresses anger and frustration towards an individual, calling them a derogatory term. The author questions whether the person's actions constitute slander and suggests the possibility of a lawsuit. They also criticize the Republican Party for their silence, blaming them for the situation and urging others to vote for Democratic candidates to protect America. The hashtags used are \#NeverForget and \#VoteBlueToSaveAmerica.

(2) The Tweet primarily focuses on criticizing Republicans (or the Republican Party/GOP). It expresses dissatisfaction with their silence and holds them responsible for a certain issue. It also includes a hashtag (\#VoteBlueToSaveAmerica) encouraging support for the Democratic Party as an alternative.

(3) The Tweet expresses aversion, dislike, distrust, blame, criticism, and negative sentiments towards Republicans (or the Republican Party/GOP). It criticizes their silence and assigns responsibility to them, using negative language to describe their actions.

(4) Yes, the Tweet expresses aversion, dislike, distrust, blame, criticism, and negative sentiments towards Republicans (or the Republican Party/GOP).
\end{ttfamily}
\end{quote}

The second tweet reads: ``Conservative stupidity will do that.'' The concept-specific breakdown for this tweet, using the same CGCoT prompts for aversion to Republicans, is the following.

\begin{quote}
\small
\begin{ttfamily}
(0) Original Tweet: ``Conservative stupidity will do that.''

(1) The tweet criticizes conservative individuals by suggesting that their stupidity is responsible for certain negative consequences.

(2) The tweet primarily focuses on conservatives, which includes Republicans, right-wingers, and individuals who support MAGA (Make America Great Again).

(3) The tweet expresses criticism and negative sentiments towards conservatives, including Republicans and individuals who align with right-wing ideologies.

(4) Yes, the tweet expresses criticism and negative sentiments towards conservatives, including Republicans.
\end{ttfamily}
\end{quote}

These concept-specific breakdowns are then pairwise compared using the approach described in Section~\ref{subsection:pairwise_comparison}. The generated pairwise comparison reads as follows.

\begin{quote}
\small
\begin{ttfamily}
Tweet Description 1 expresses greater aversion, dislike, distrust, blame, criticism, or negative sentiments of Republicans compared to Tweet Description 2.
\end{ttfamily}
\end{quote}

\section{Application Results}
\subsection{Face Validity of the Aversion CGCoT Pairwise Scores}
The aversion CGCoT pairwise scores show face validity when examining the rank ordering of tweets. We showcase three tweet examples with different aversion scores. The first has the highest score, the second has a score closest to the mean of the measure, and the third has the lowest score. We first examine the tweets that have these specific aversion to Republicans CGCoT pairwise scores.

\begin{itemize}
    \item Highest: ``\#ArrestTrump and all associates immediately. Try them for treason. PUBLIC CAPITAL PUNISHMENT. They're all traitors \& murderers. NEVER FORGET all the blood they have on their hands \#Trump \#Republicans \#1776RestorationMovement \#January6thCommittee \#Jan6Hearings''
    \item Middling: ``@newsmax America doesn't have to worry about socialism. The threat to American democracy is not socialism, it is the radicalized and unhinged Republicans.''
    \item Lowest: ``House Republican leader Kevin McCarthy calls the situation at the southern border a humanitarian and national security crisis.''
\end{itemize}

The ordering of these tweets is intuitively consistent with our concept of interest. The intensity of aversion expressed in tweets decreases from the highest to the lowest CGCoT pairwise scores. The middling tweet expresses aversion to Republicans but is not nearly as intense as the tweet with the highest CGCoT pairwise score.

We then examine the tweets associated with the same ordering of the aversion to Democrats CGCoT pairwise scores.

\begin{itemize}
    \item Highest: ``Very true. I keep wondering the same. The Demonrats have sunk deeper into the swamp! They've been exposed, but that's only let them do their dirty deeds defiantly out in the open. Why has no one been stopped or paid for their crimes?''
    \item Middling: ``I kind of like blocking the liberals and only seeing common sense comments.''
    \item Lowest: ``Get over it, scoop! It's obvious he's waiting to put it all together with the DNC this week. Then I bet he'll be holding pressers with his policies and all. But, this incessant reporter whining is for the birds.''
\end{itemize}

There is a similar pattern in the intensity of the aversion expressed in the tweets over the CGCoT pairwise scores. The middling tweet is a jab at liberals, but it is not the same type of criticism leveled at Democrats as the tweet with the highest CGCoT pairwise score. We also note that ``Demonrats'' is correctly identified as an intense insult to Democrats, an example of the potential contextualizing capabilities of generative LLMs. 

\subsection{Comparing CGCoT Pairwise Scores with Human Coders}
\label{sec:comparing_cgcot_human_coders}
We compare both scores with the number of human coders that labeled a tweet as containing aversion to Republicans or aversion to Democrats. Despite the differences in granularity between the discrete labels assigned by human coders and the continuous measures estimated using our proposed approach, we can gain insights into the overall prevalence and intensity of aversion expressed in the tweets by comparing the number of coders labeling each tweet as containing aversion to the distribution of tweets along the estimated scores.

\begin{figure}[!htbp]
    \centerline{\includegraphics[width=0.5\textwidth]{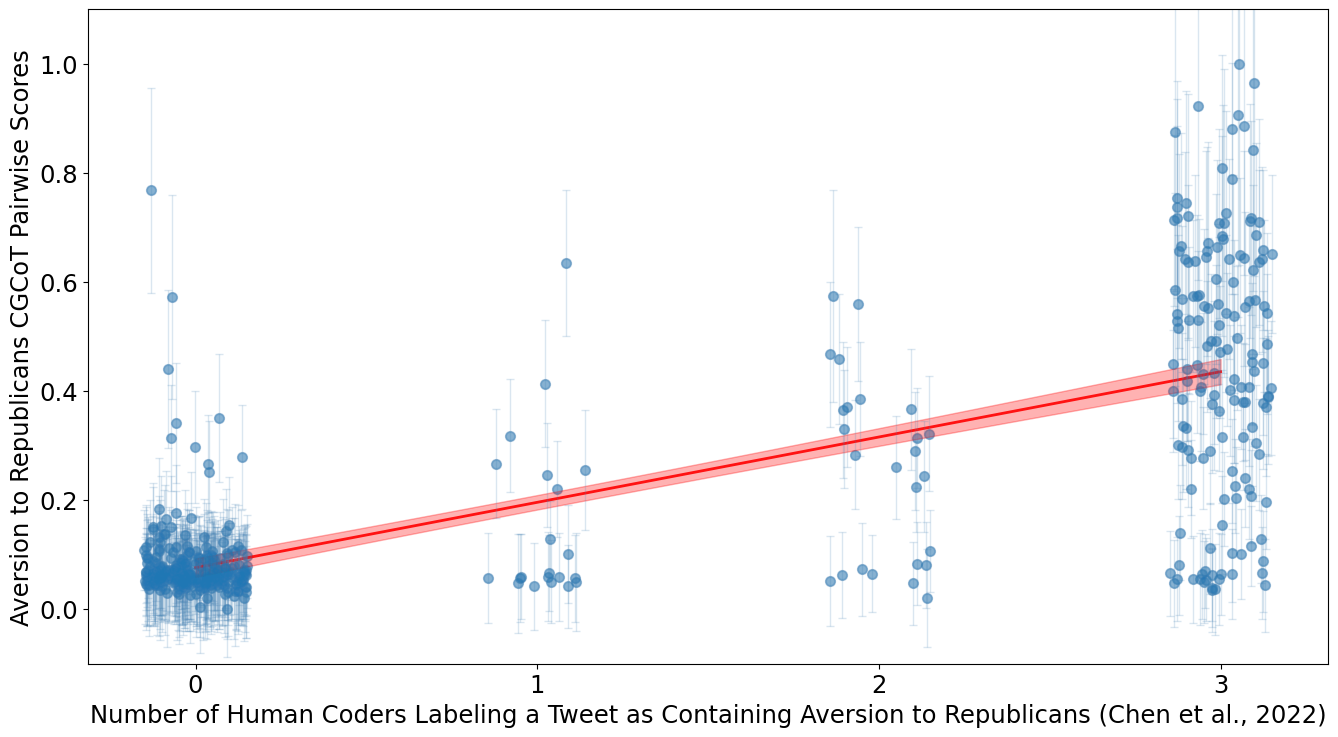}}
    \caption{Aversion to Republicans CGCoT pairwise scores are strongly associated with the number of coders applying an aversion to Republicans label. CGCoT pairwise score estimates are shown with 95\% confidence intervals derived from quasi-standard errors. A linear regression line is drawn through the points with a 95\% confidence band.}
    \label{fig:aversionreps_humancoders}
\end{figure}

We first compare the aversion to Republicans CGCoT pairwise scores against the number of human coders that labeled a tweet as containing aversion to Republicans (Figure~\ref{fig:aversionreps_humancoders}). We find a positive correlation between these two measures. We note that the wider distribution of CGCoT pairwise scores for tweets with two or three coders labeling a tweet as containing aversion to Republicans is a function of the pairwise comparisons: tweets that do not contain any aversion to Republicans tend to tie with each other in matchups, resulting in scores ``clumping.'' In the Appendix, we examine the tweets with the lowest aversion to Republicans CGCoT pairwise scores among tweets labeled by three coders as containing aversion to Republicans.

We repeat this exercise for aversion to Democrats. We also find a positive correlation when comparing aversion to Democrats CGCoT pairwise scores against the number of human coders that labeled a tweet as containing aversion to Democrats. Figure~\ref{fig:aversiondems_humancoders} illustrates this comparison. In the Appendix, we again examine the tweets with the lowest aversion to Democrats CGCoT pairwise scores among tweets labeled by three coders as containing aversion to Democrats.

\begin{figure}[!htbp]
    \centerline{\includegraphics[width=0.5\textwidth]{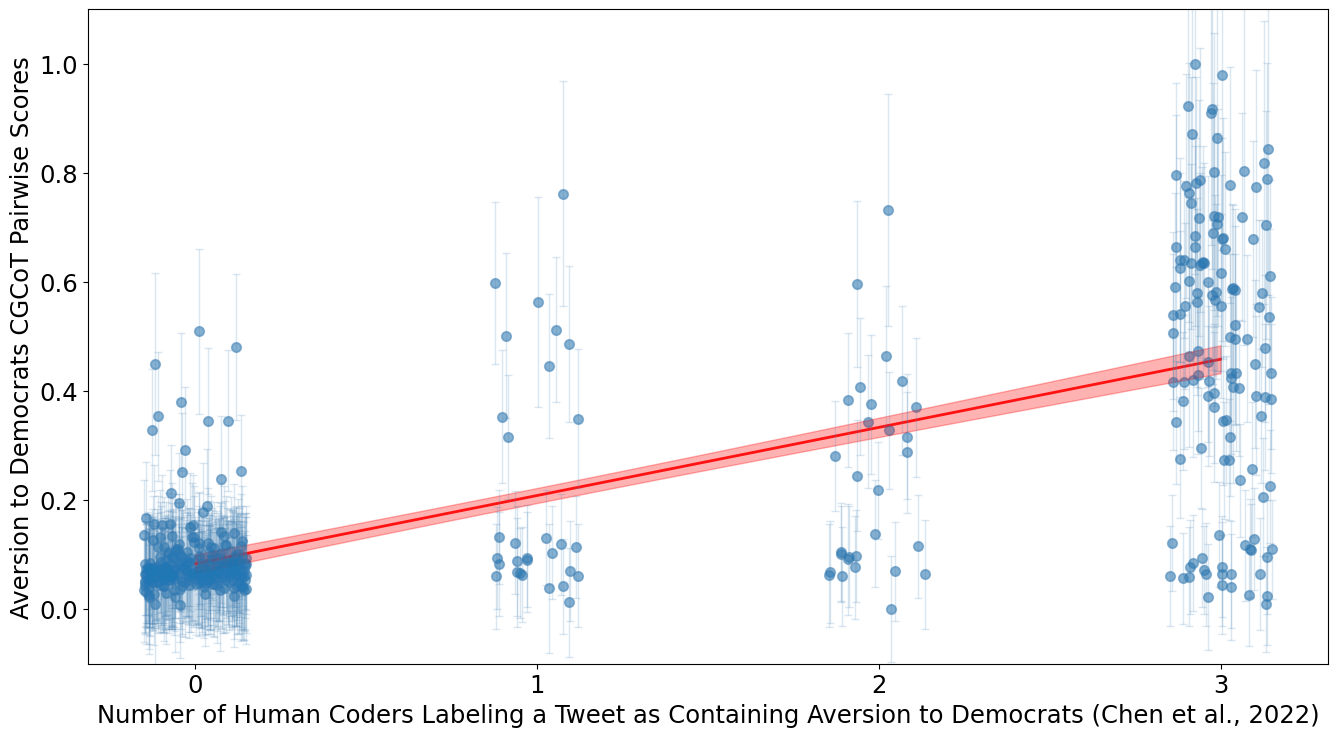}}
    \caption{Aversion to Democrats CGCoT pairwise scores are strongly associated with the number of coders applying an aversion to Democrats label. CGCoT pairwise score estimates are shown with 95\% confidence intervals derived from quasi-standard errors. A linear regression line is drawn through the points with a 95\% confidence band.}
    \label{fig:aversiondems_humancoders}
\end{figure}

We also calculate Spearman's rank correlations between human labels and an array of text scoring methods, including CGCoT pairwise scores. We do this for both types of aversion using (1) Wordfish with the tweets only; (2) Wordfish with the tweets' concept-specific breakdowns; (3) the pairwise comparison approach with GPT-3.5 using the tweets only (``non-CGCoT tweets-only pairwise scores''); and (4) CGCoT pairwise scores. 

Wordfish is one of the most popular unsupervised text scoring methods in the social and political sciences and has been used in many recent works \cite{bailey2023measuring}. The primary goal of Wordfish \cite{wordfish_slapin_proksch} is to estimate the position of a document along a single dimension. The assumption is that the rate at which tweet $i$ mentions word $j$ is drawn from a Poisson distribution. The functional form of the model is 
\[y_{ij} \sim Poisson(\lambda_{ij})\]
\[\lambda_{ij} = \exp\left(\alpha_{i} + \psi_j + \beta_j \omega_{i}\right)\]
where $y_{ij}$ is the count of word $j$ in tweet $i$, $\alpha$ is the set of tweet fixed effects, $\psi$ is the set of word fixed effects, $\beta$ is an estimate of a word-specific weight that reflects the importance of word $j$ in discriminating between positions, and $\omega_i$ is tweet $i$'s position. We fit this model using the \texttt{quanteda} R package \cite{quanteda}. We used standard preprocessing steps: we removed symbols, numbers, and punctuation. We also stemmed the words. Lastly, we imposed minimum word counts and word-document frequency counts to prevent non-convergence. For aversion to Republicans, a word had to be used at least 4 times across 4 or more tweets. For aversion to Democrats, a word had to be used at least 6 times across 6 or more tweets. We used these configurations when using Wordfish with both the tweets and the concept-specific breakdowns. 

\begin{table}[!ht]
\renewcommand{\arraystretch}{1.3}
\caption{CGCoT pairwise scores outperform other unsupervised text scoring methods based on Spearman's rank correlation coefficient with the number of human coders labeling a tweet as containing aversion to Republicans (``GOP'') and Democrats (``Dems.'').}
\centering
\begin{tabular}{l|cc}
\hline
                                          & \multicolumn{2}{c}{\bf Aversion to...} \\
                                          & \textbf{GOP}   & \textbf{Dems.}         \\ \hline
Wordfish w/ Tweets Only                   & 0.03                   & 0.04              \\
Wordfish w/ Concept-Specific Breakdowns   & 0.55                   & 0.22              \\
Non-CGCoT Tweets-Only Pairwise Scores     & 0.55                   & 0.56              \\
CGCoT Pairwise Scores                     & 0.64                   & 0.61             
\end{tabular}
\label{tab:correlations_aversion}
\end{table}

Table~\ref{tab:correlations_aversion} shows these correlations for both aversions to Republicans and Democrats. The results demonstrate the utility of both CGCoT and pairwise comparisons, with notable gains in correlation when moving from the use of tweets to concept-specific breakdowns of the tweets, and from Wordfish to pairwise comparisons. Our proposed procedure of using CGCoT with LLM pairwise comparisons yields a measure that most closely aligns with human judgments compared to other text scoring approaches. 

\subsection{CGCoT Pairwise Scores are Competitive with Supervised Learning Approaches}
To further analyze the validity of the aversion CGCoT pairwise scores, we create binary labels using cutoffs in the two aversion measures. For each measure, we label all tweets with CGCoT pairwise scores above the mean of the CGCoT pairwise scores as 1 (expresses aversion to a specific party) and all tweets with CGCoT pairwise scores below the mean as 0 (does not express aversion to a specific party). While binarizing the CGCoT pairwise scores by labeling observations above the mean as 1 and those below the mean as 0 is slightly arbitrary, this approach is also guided by a principled decision to use the central tendency of the measure as the threshold. Future work will use a training set to choose a more accurate cutoff, which would almost certainly improve results. We repeat this process for the measure generated using GPT-3.5 pairwise comparisons of the tweets only (i.e., non-CGCoT tweets-only comparisons). We also compare CGCoT pairwise scores with a RoBERTa-Large model fine-tuned using \cite{affective_polarization_csmap}'s training set of 3,000 hand-labeled political tweets with hyperparameters chosen using the validation set. Table~\ref{tab:aversion_clf} contains the performance metrics of the two cutoff classifiers and the RoBERTa-Large model.

\begin{table*}[!htbp]
\renewcommand{\arraystretch}{1.3}
\caption{The F1 score, precision, and recall of the non-CGCoT tweets-only pairwise scores cutoff classifier, the RoBERTa-Large classifier fine-tuned on \cite{affective_polarization_csmap}'s training set, and the CGCoT pairwise scores cutoff classifier.}
\centering
\begin{tabular}{l|l|ccc}
\hline
\textbf{Classifier}                                            & \textbf{Aversion to}      & \textbf{F1} & \textbf{Precision} & \textbf{Recall} \\ \hline
\multirow{2}{*}{Tweets-Only Pairwise Scores Cutoff Classifier}       & Republicans               & 0.70        & 0.64           & 0.76            \\
                                                               & Democrats                 & 0.67        & 0.58           & 0.79            \\ \hline
\multirow{2}{*}{Fine-Tuned RoBERTa-Large Model}                & Republicans               & 0.81        & 0.82           & 0.80            \\
                                                               & Democrats                 & 0.81        & 0.82           & 0.81            \\ \hline
\multirow{2}{*}{CGCoT Pairwise Scores Cutoff Classifier}             & Republicans               & 0.84        & 0.89           & 0.79            \\
                                                               & Democrats                 & 0.79        & 0.84           & 0.75           
\end{tabular}
\label{tab:aversion_clf}
\end{table*}

The performance metrics show that CGCoT pairwise scores outperform non-CGCoT tweets-only pairwise scores on all metrics except for recall for aversion to Democrats. The metrics also show that the aversion to Republicans CGCoT pairwise scores outperform the fine-tuned RoBERTa-Large model on F1 and precision and are nearly equivalent on recall. Similarly, the aversion to Democrats CGCoT pairwise scores are comparable with the fine-tuned RoBERTa-Large model on F1. The former performs better on precision, and the latter performs better on recall. Again, the CGCoT pairwise comparison cutoff classifier's predictions were calculated using no hand-labeled tweets, except for a small set of 200 hand-labeled pilot tweets used to develop the CGCoT prompts for the concept-specific breakdowns. In other words, our approach drastically reduces the need for training coders and hand-labeling data while still retaining the expertise needed to analyze this nuanced and complex concept expressed in social media posts.

\section{Conclusion}
We develop a novel text scoring framework that leverages pairwise comparisons and a prompting procedure called concept-guided chain-of-thought (CGCoT), which creates concept-specific breakdowns of the texts. We then prompt GPT-3.5 to make pairwise decisions between the concept-specific breakdowns along a latent concept. We call the resulting measures CGCoT pairwise scores. We apply the approach to better understand affective polarization on social media and derive two novel latent measures of aversion to specific political parties in tweets. 

We find that the measures largely correlate with how humans interpret aversion to Republicans and aversion to Democrats on Twitter. We also find that using \textit{both} CGCoT and pairwise comparisons with LLMs is crucial, as scores that do not use one or both of these techniques are demonstrably worse. Moreover, our approach can estimate scores that are competitive with or outperform both unsupervised and supervised approaches. We show that using a cutoff with the score yields binary classifications that are highly competitive with a RoBERTa-Large model fine-tuned on \textit{thousands} of human-labeled tweets. Our findings suggest that using substantive knowledge with generative LLMs can not only be useful for calculating high-quality continuous measures but can also be useful for generating discrete classifications with high performance with the use of very little human-labeled data. 

Our findings align with the notion that the LLM synthesizes information about complex concepts such as affective polarization, allowing it to reliably and coherently evaluate latent constructs, abstract concepts, stances, and sentiments within texts using its pattern recognition capabilities. While we provide information about what constitutes a ``Republican'' and ``Democrat'' being targeted in our CGCoT prompts, we still assume that the LLM is able to identify Republican or Democratic figures and organizations, such as Donald Trump, Joe Biden, and the DCCC. Additionally, we assume that the LLM possesses the capability to recognize aversion within the presented texts. Again, this capability stems from the presence of many instances of political contention in social media and other forms of content. However, the precise impact of this training on both CGCoT and the pairwise comparisons remains obscured due to the black box nature of GPT and requires further investigation. 

It is also well-known that there is a significant mental toll on people identifying attacks against individuals/groups and harmful content for data labeling and content moderation purposes. Our approach, which rivals the binary prediction performance of language models fine-tuned on thousands of hand-labeled social media posts, can help avoid having human coders label thousands of posts containing potentially harmful or sensitive content. 

There are still many open questions around this framework, as well as future directions of work. We apply the approach using only one generative LLM to one substantive problem. We have also not yet analyzed the consistency of pairwise comparisons over repeated promptings and how sensitive the pairwise comparisons are to the prompt's wording. We have also not yet considered how the timing of social media posts comports with the LLM's training data. Ongoing work aims to answer many of these open questions, including expanding the framework to use recently developed techniques such as retrieval-augmented generation \cite{RAG_Lewis_etal_2020} and studying how the outcomes of pairwise comparisons may change when using semantically equivalent prompts. Despite the approach's current limitations, it estimates scores that agree with human judgments of the texts along different dimensions of interest. It lends a better understanding of how human-AI collaboration can be used to improve the quantification and measurement of complex latent concepts.

\section*{Acknowledgements}
We gratefully acknowledge that the Center for Social Media and Politics at New York University is supported by funding from the John S. and James L. Knight Foundation, the Charles Koch Foundation, Craig Newmark Philanthropies, the William and Flora Hewlett Foundation, and the Siegel Family Endowment. This work was supported in part through the NYU IT High Performance Computing resources, services, and staff expertise.

{
\bibliographystyle{IEEEtran}
\bibliography{bib}
}

\appendix
\subsection{Correlations Between Scores with Differing Number of Matchups}
The reported results are from 20 matchups per tweet ID, for a total of 10,000 matchups. We analyze how the CGCoT pairwise scores calculated using 5, 10, 15, and 20 matchups per tweet ID correlate for both measures. Results are in Table~\ref{tab:correlation_matchups_SI_aversion_agaisnt_republicans} and Table~\ref{tab:correlation_matchups_SI_aversion_against_democrats}. Across all configurations, correlations are greater than 0.90. 

\begin{table}[!htbp]
\renewcommand{\arraystretch}{1.3}
\caption{Pearson correlations between CGCoT pairwise scores estimated using 5, 10, 15, and 20 matchups per tweet ID for aversion to Republicans.}
\centering
\begin{tabular}{c|cccc}
   & 5     & 10    & 15    & 20    \\ \hline
5  & 1.000 & 0.937 & 0.940 & 0.953 \\
10 & 0.937 & 1.000 & 0.964 & 0.979 \\
15 & 0.940 & 0.964 & 1.000 & 0.986 \\
20 & 0.953 & 0.979 & 0.986 & 1.000
\end{tabular}
\label{tab:correlation_matchups_SI_aversion_agaisnt_republicans}
\end{table}

\begin{table}[!htbp]
\renewcommand{\arraystretch}{1.3}
\caption{Pearson correlations between CGCoT pairwise scores estimated using 5, 10, 15, and 20 matchups per tweet ID for aversion to Democrats.}
\centering
\begin{tabular}{c|cccc}
   & 5     & 10    & 15    & 20    \\ \hline
5  & 1.000 & 0.934 & 0.929 & 0.943 \\
10 & 0.934 & 1.000 & 0.958 & 0.976 \\
15 & 0.929 & 0.958 & 1.000 & 0.984 \\
20 & 0.943 & 0.976 & 0.984 & 1.000
\end{tabular}
\label{tab:correlation_matchups_SI_aversion_against_democrats}
\end{table}

\subsection{Tweets with the Lowest CGCoT Pairwise Scores that were Labeled by Three Coders as Containing Aversion}
\subsubsection{Aversion to Republicans}
Looking at just the tweets labeled by three coders as containing aversion to Republicans, we examined the three tweets with the lowest aversion to Republicans CGCoT pairwise scores. The text of these tweets is as follows. 
\begin{enumerate}
    \item ``Dear @POTUS: Are those forgotten men \& women who you say never protest the same people showing up at statehouses armed w/military-style weaponry? Are those same forgotten ones the same who call themselves Boogaloo? Or are they those very fine Nazis you favor? All of the above?''
    \item ``Cadet bone-spur \& tribe be innocent then they should welcome investigations clearing their good name besmirched by furtive conniving fake news liberals.''
    \item ``Trump campaign thought their `huge news' on pre-existing conditions had Democrats cornered -- but it backfired spectacularly https://t.co/Ue8ugrCzRF''
\end{enumerate}
GPT-3.5 makes mistakes in the interpretation of certain phrases in two of these tweets: it misinterprets a message addressed to the @POTUS account as directed towards President Biden, not President Trump, and it did not recognize ``Cadet bone-spur'' to be a derisive nickname for Trump. In the third tweet, GPT interpreted a vague headline describing something backfiring against the Trump campaign as not expressing aversion to Republicans, an arguably correct interpretation.

\subsubsection{Aversion to Democrats}
Looking at just the tweets labeled by three coders as containing aversion to Democrats, we again examine the three tweets with the lowest aversion to Democrats CGCoT pairwise scores. The text of these tweets is as follows.  
\begin{enumerate}
    \item ``Maybe if Yang and Tulsi were running the party, I might have a much better opinion of the Democratic Party.  However as it stands, I cannot.  I do respect Yang for wanting to help fighters get paid fairly.  My brother boxed for 20 years and doesnt have anything to show for it.''
    \item ``This is what the MSM and libs won't ever tell you. And folks reply to this with all sorts of whataboutism, that it's perfectly fine these cops were injured by violent protesters. So that means all Jan 6 defendants should never have been arrested? AmIright?''
    \item ``Regular law enforcement like any other regulated profession so that bad cops can't just get rehired at the next department over.' Wtf does this mean? Proof your statements, Dems.''
\end{enumerate}
Here, GPT-3.5 interpreted the text differently from humans. For example, in the first tweet, it did not interpret someone describing how they do not respect the Democratic Party as expressing aversion to Democrats. In the second tweet, GPT-3.5 did not interpret ``libs'' as an insult towards liberals. Lastly, in the third tweet, the author asked Democrats to ``proof your statements,'' which GPT-3.5 did not interpret as an insult or criticism of Democrats.

\end{document}